\documentclass[runningheads]{llncs}
\usepackage[T1]{fontenc}
\usepackage{graphicx}
\usepackage{orcidlink}
\usepackage{caption}
\usepackage{subcaption}

\usepackage{hyperref}
\usepackage{booktabs}
\usepackage{multirow}
\usepackage{subcaption}
\usepackage{amsmath}
\usepackage{amssymb}
\usepackage{cleveref}
\crefname{figure}{Fig.}{Figs.}
\Crefname{figure}{Fig.}{Figs.}
\usepackage[symbol]{footmisc}

\usepackage[utf8]{inputenc}
\usepackage{tikz}
\usepackage{pgfplots}
\usepgfplotslibrary{groupplots,dateplot}
\usetikzlibrary{patterns,shapes.arrows}
\pgfplotsset{compat=newest}
\def\myplotmark#1{%
  \begin{pgfpicture}\pgfuseplotmark{#1}\end{pgfpicture}%
}

\begin{document}
\title{Few-Shot Segmentation of Historical Maps via Linear Probing of Vision Foundation Models}
\titlerunning{Few-Shot Segmentation of Historical Maps Leveraging Foundation Models}
\author{Rafael Sterzinger\orcidlink{0009-0001-0029-8463} \and
Marco Peer \orcidlink{0000-0001-6843-0830} \and
Robert Sablatnig\orcidlink{0000-0003-4195-1593}}
\authorrunning{R. Sterzinger et al.}
\institute{Computer Vision Lab, TU Wien, Vienna, AUT
\email{\{firstname.lastname\}@tuwien.ac.at}}
\maketitle              
\begin{abstract}
As rich sources of history, maps provide crucial insights into historical changes, yet their diverse visual representations and limited annotated data pose significant challenges for automated processing.
We propose a simple yet effective approach for few-shot segmentation of historical maps, leveraging the rich semantic embeddings of large vision foundation models combined with parameter-efficient fine-tuning.
Our method outperforms the state-of-the-art on the Siegfried benchmark dataset in vineyard and railway segmentation, achieving +5\% and +13\% relative improvements in mIoU in 10-shot scenarios and around +20\% in the more challenging 5-shot setting.
Additionally, it demonstrates strong performance on the ICDAR 2021 competition dataset, attaining a mean PQ of 67.3\% for building block segmentation, despite not being optimized for this shape-sensitive metric, underscoring its generalizability.
Notably, our approach maintains high performance even in extremely low-data regimes (10- \& 5-shot), while requiring only 689k trainable parameters -- just 0.21\% of the total model size.
Our approach enables precise segmentation of diverse historical maps while drastically reducing the need for manual annotations, advancing automated processing and analysis in the field.
Our implementation is publicly available at: \href{https://github.com/RafaelSterzinger/few-shot-map-segmentation}{https://github.com/RafaelSterzinger/few-shot-map-segmentation}.
\end{abstract}

\begin{figure}[t]
\centering
\begin{subfigure}{0.24\textwidth}
\includegraphics[width=\textwidth]{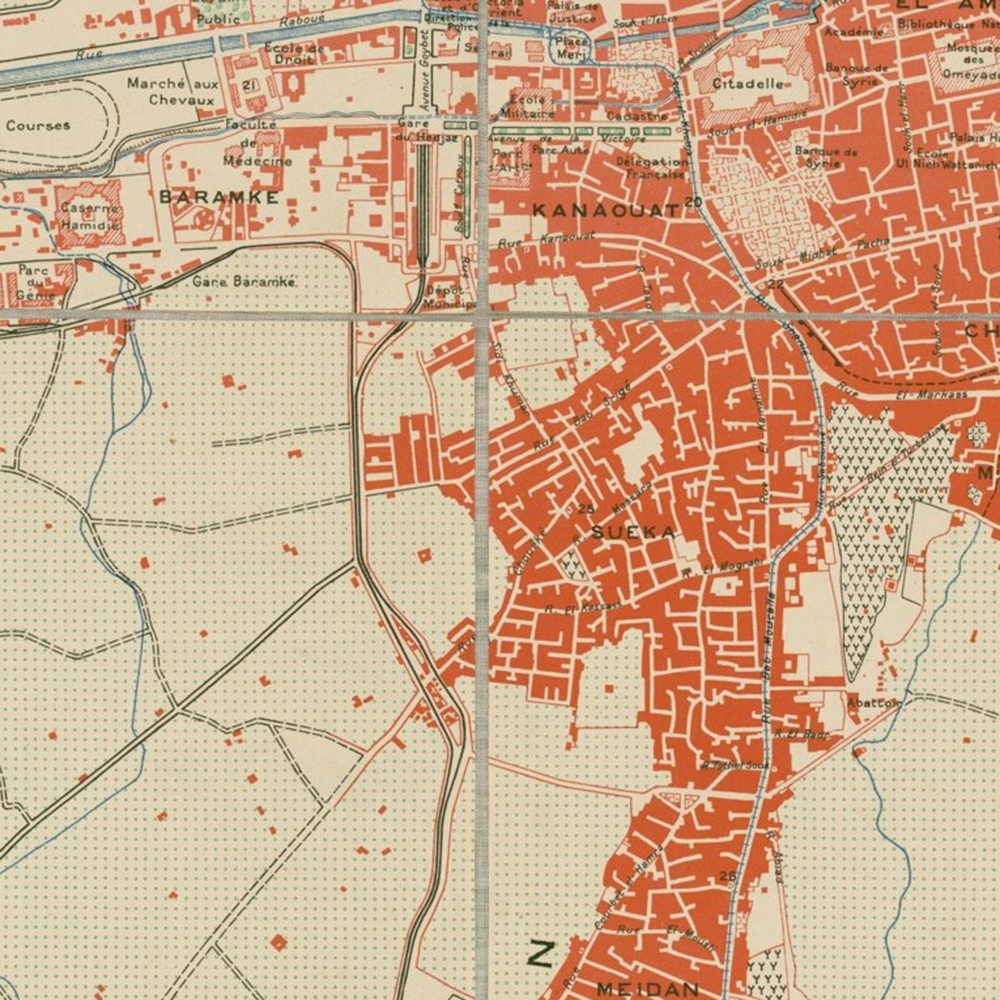}
\end{subfigure}
\begin{subfigure}{0.24\textwidth}
\includegraphics[width=\textwidth]{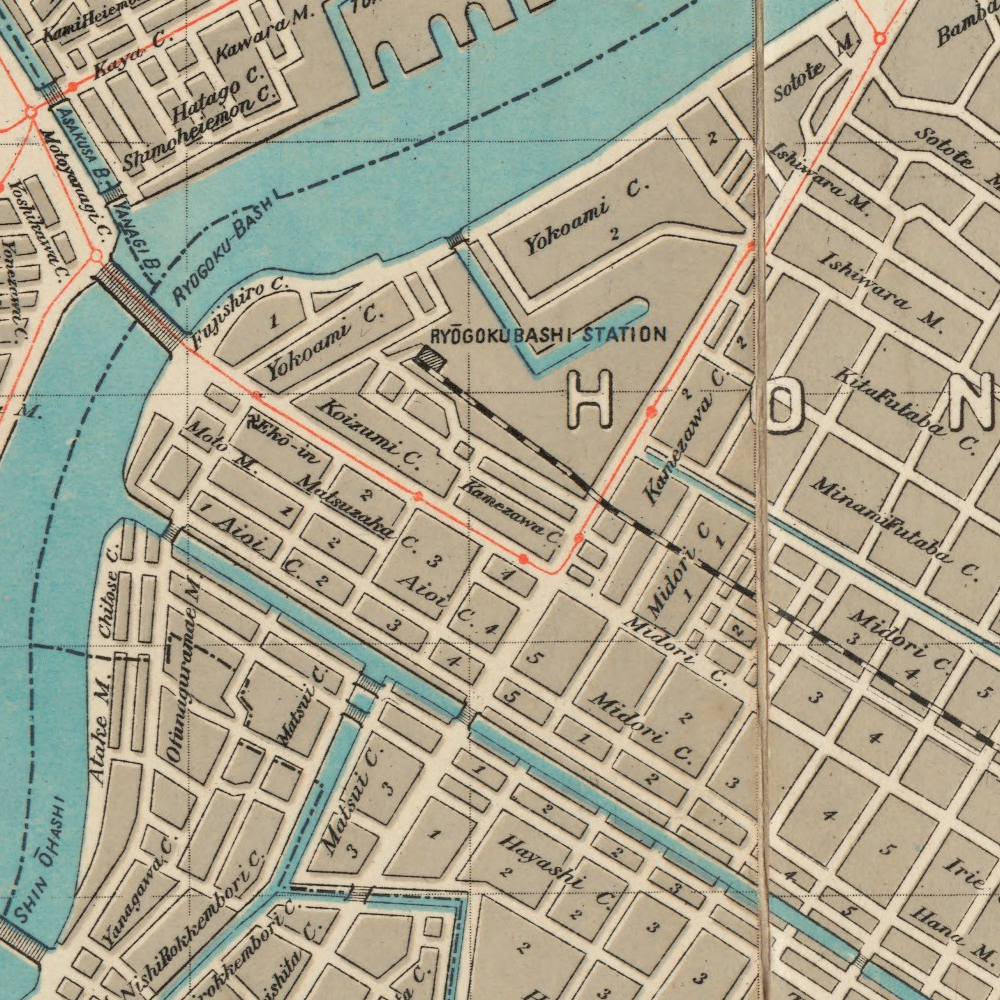}
\end{subfigure}
\begin{subfigure}{0.24\textwidth}
\includegraphics[width=\textwidth]{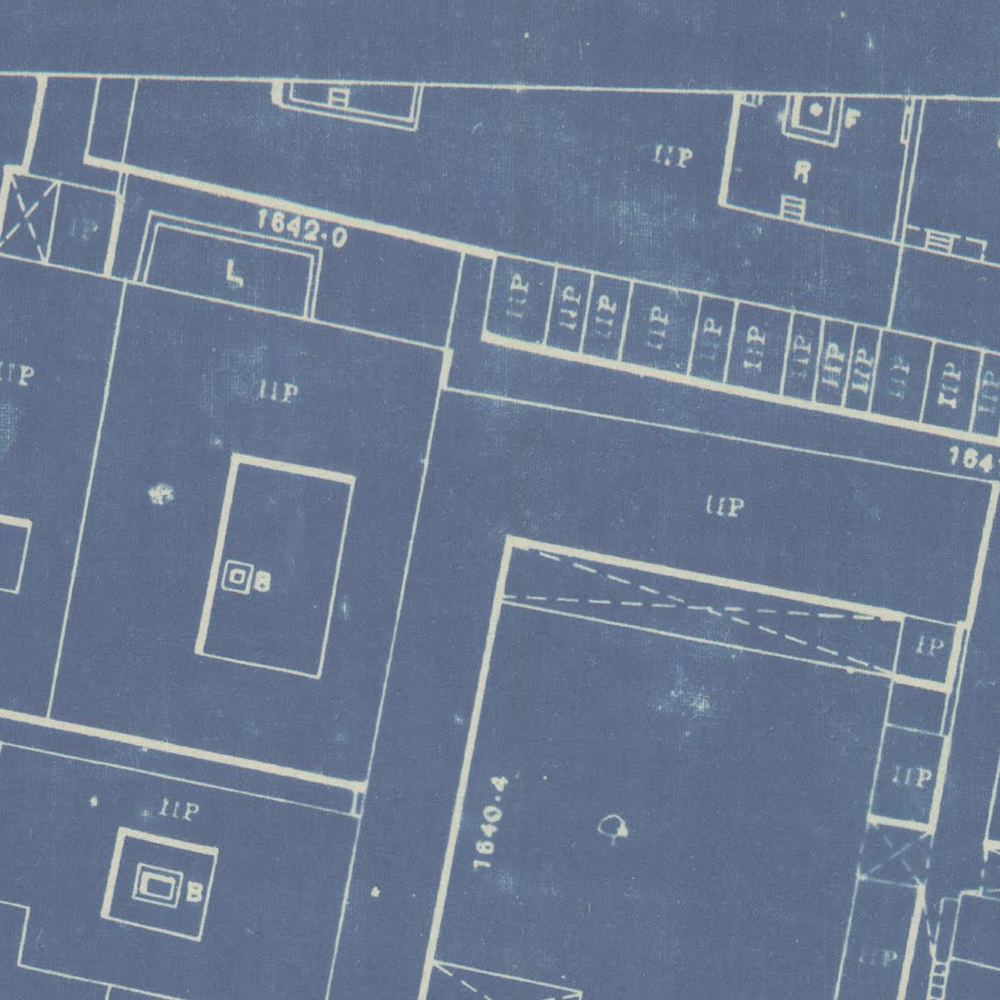}
\end{subfigure}
\begin{subfigure}{0.24\textwidth}
\includegraphics[width=\textwidth]{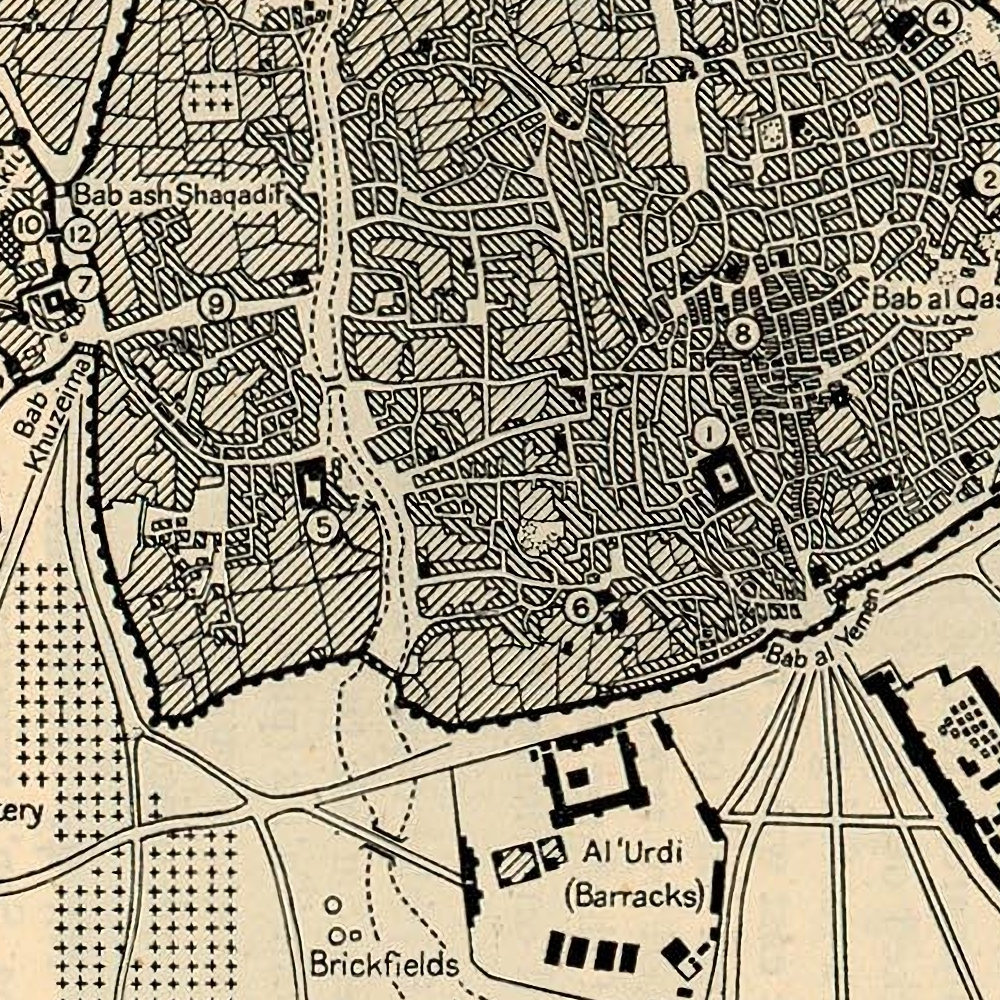}
\end{subfigure}

%\vspace{1pt}
%\begin{subfigure}{0.24\textwidth}
%\includegraphics[width=\textwidth]{figs/map_teaser/3788255894.png}
%\end{subfigure}
%\begin{subfigure}{0.24\textwidth}
%\includegraphics[width=\textwidth]{figs/map_teaser/3788258904.png}
%\end{subfigure}
%\begin{subfigure}{0.24\textwidth}
%\includegraphics[width=\textwidth]{figs/map_teaser/3788304369.png}
%\end{subfigure}
%\begin{subfigure}{0.24\textwidth}
%\includegraphics[width=\textwidth]{figs/map_teaser/3788418516.png}
%\end{subfigure}
\caption{An excerpt of historical city maps from around the world, published between 1720 and 1950.~\copyright Petitpierre et al.}
\label{fig:map_varity}
\end{figure}

\section{Introduction}
As invaluable records of the past, historical maps provide critical insights into geographic, infrastructural, and sociopolitical changes.
However, their automated analysis remains a significant challenge due to the wide stylistic variability, inconsistent annotations, and frequent physical degradation inherent in these documents~\cite{petitpierre_generic_2021,xia_mapsam_2024}.
In contrast to modern maps, which often conform to standardized cartographic norms, historical maps are profoundly heterogeneous, both in visual appearance and semantic content, as illustrated in Figure~\ref{fig:map_varity}.

As a result, models struggle to generalize effectively due to the domain gap between maps, making it difficult to transfer knowledge from one corpus to another~\cite{petitpierre_generic_2021}.
Another key obstacle in developing robust models for automated historical map segmentation is the scarcity of labeled data, a substantial problem that extends beyond historical maps: as annotating historical artifacts demands expert knowledge and considerable manual effort, which renders the task cost-ineffective, it is impractical to build the large training corpora typically required for supervised learning~\cite{sterzinger_drawing_2024,sterzinger_fusing_2024}.

\begin{figure}[t]
\centering
\begin{subfigure}{0.35\textwidth}
\includegraphics[width=\textwidth]{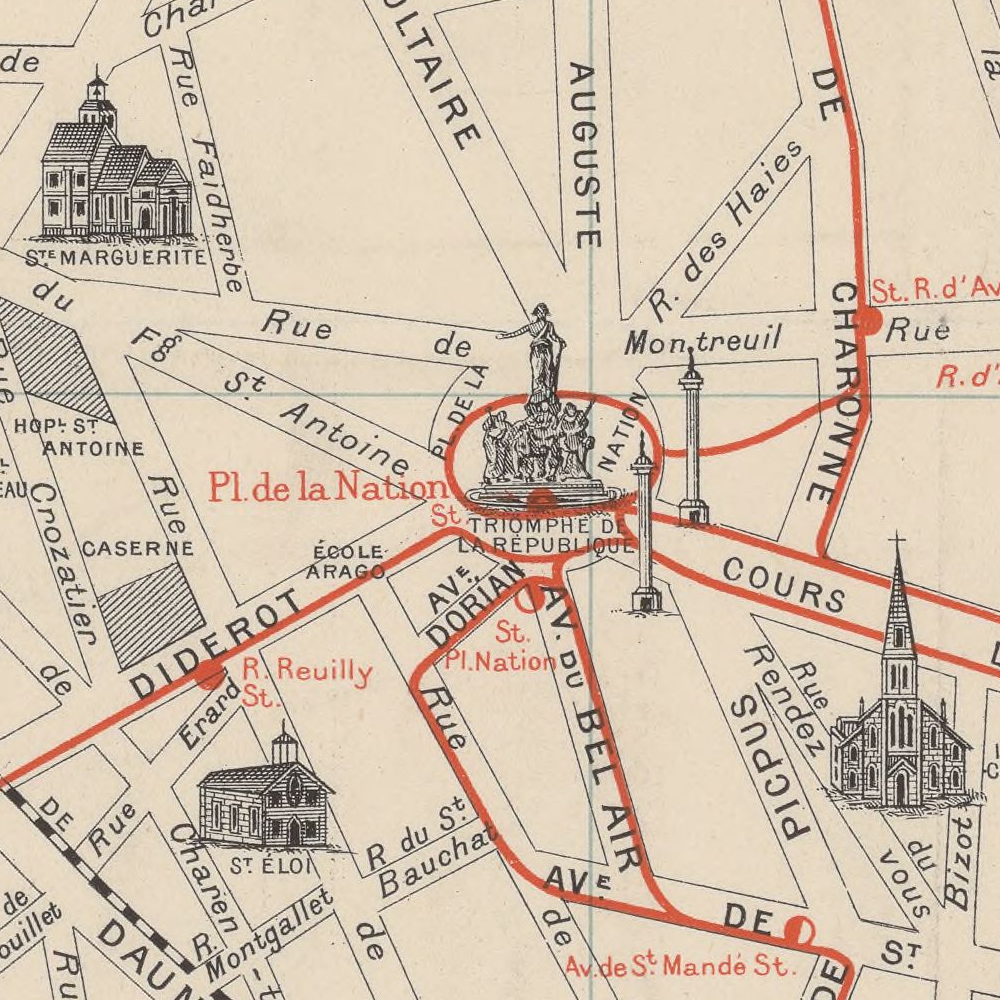}
\subcaption{Original}
\end{subfigure}
\hspace{1cm}
\begin{subfigure}{0.35\textwidth}
\includegraphics[width=\textwidth]{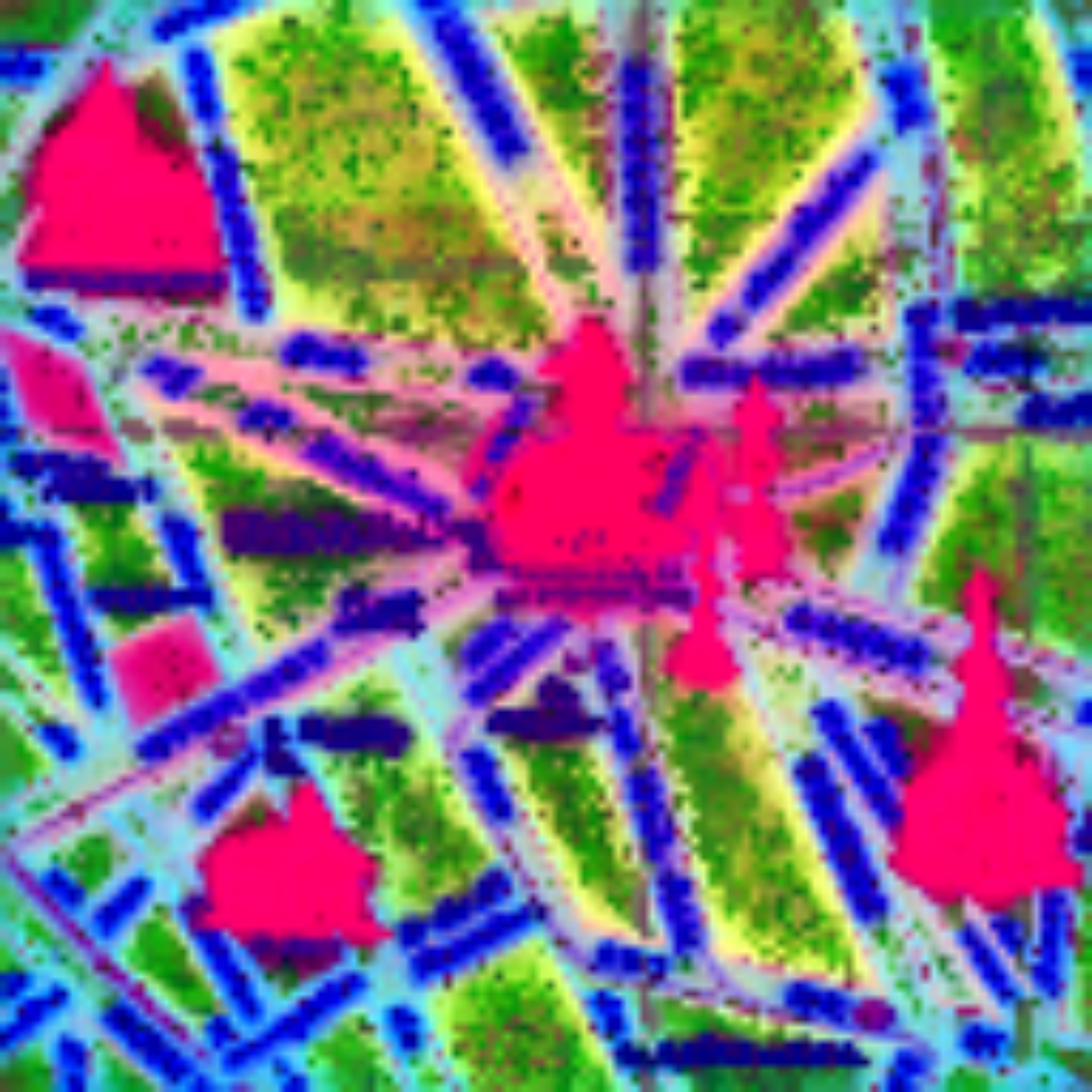}
\subcaption{PCA Radio}
\end{subfigure}
\caption{An illustration of the first three principal components of RADIO-H~\cite{ranzinger_am-radio_2024} feature embeddings of a map of Paris: despite no prior training in this specialized domain, meaningful classes have already emerged: landmarks, building blocks, streets, and street names are clearly distinguishable.}
\label{fig:pca_radio}
\end{figure}

As a consequence of these challenges, historical map segmentation presents a promising, yet largely untapped, application for few-shot learning, a paradigm which enables models to learn from limited labeled data by leveraging pretrained representations of large vision foundation models~\cite{snell_prototypical_2017,ding_self-regularized_2023,he_apseg_2024,sun_vrp-sam_2024}.
Over the years, numerous vision foundation models have emerged that are trained on vast corpora of natural images to serve as powerful backbones for various downstream tasks across diverse domains~\cite{ranzinger_am-radio_2024}. 
When processing images, these models can generate rich features, which are particularly beneficial for dense prediction tasks such as segmentation, sometimes even for domains such as historical maps.
As illustrated in Figure~\ref{fig:pca_radio}, our visualization shows the first three principal components of embeddings obtained from RADIO~\cite{ranzinger_am-radio_2024}, which already capture meaningful semantic structures, offering a strong basis for downstream tasks.

Ideally, these expressive representations can bridge the gap across stylistic differences and deliver satisfactory results even in low-data regimes.
However, while often pretrained on natural images, historical maps present a unique challenge: their structural complexity means features such as roads, railways, and river patterns can appear visually similar, although semantically different.
Based on this observation, we necessitate that the models need to learn the abstract distinctions between these elements, highlighting the crucial need for model adaptation, for example, through parameter-efficient fine-tuning~\cite{xia_mapsam_2024}.

Given these challenges -- the heterogeneity of historical maps, the scarcity of annotated data, and the domain gap between natural images and historical documents -- Xia et al.~\cite{xia_mapsam_2024} explored leveraging SAM~\cite{kirillov_segment_2023} and its representations for improved few-shot map segmentation.
We extend their work and investigate other types of foundation models such as DINOv2~\cite{oquab_dinov2_2024} or RADIO~\cite{ranzinger_am-radio_2024}, which have fundamentally different training goals that yield richer and more general spatial representations.
 
Although our approach builds on MapSAM by Xia et al.~\cite{xia_mapsam_2024}, such as leveraging vision foundation models and parameter-efficient fine-tuning, its strength lies in its reduced complexity.
Unlike MapSAM, which retrains the full SAM decoder, we use a lightweight linear classifier -- yet achieve significantly better results.
Our approach exemplifies \textit{Occam's Razor}: when a simpler method outperforms more complex ones, added complexity becomes not only unnecessary but even detrimental, as more parameters increase the risk of overfitting and reduce generalization. 

In summary, our contributions include:

\begin{itemize}
    \item We provide the first comprehensive evaluation of three SOTA vision foundation models for historical map segmentation, delivering both qualitative visualizations and quantitative analysis of their feature embeddings' effectiveness in this specialized domain.
    
    \item We demonstrate that combining low-rank adaptation with linear probing of foundation models yields exceptional segmentation performance while maintaining parameter efficiency. Our extensive ablation studies validate this approach as both computationally lightweight and highly effective for historical document analysis.
    
    \item We perform an extensive evaluation on two datasets, the Siegfried and the ICDAR 2021 competition datasets. On the Siegfried dataset, we substantially outperform in all few-shot settings, especially in the challenging 5-shot setting with relative improvements of around 20\% in IoU, while requiring training of only 689k parameters (a mere 0.21\% of the overall model size).
    
    \item To facilitate future research and practical applications in historical map digitization, we release our code at: \href{https://github.com/RafaelSterzinger/few-shot-map-segmentation}{https://github.com/RafaelSterzinger/few-shot-map-segmentation}.
\end{itemize}

\section{Related Work}

Motivated by the limited generalization of models trained on homogeneous or single-map datasets, Petitpierre et al.~\cite{petitpierre_generic_2021} explore training on two stylistically diverse corpora: Paris city maps and global city maps. By evaluating CNN-based segmentation performance, they show that neural networks trained on large, diverse corpora can integrate abstract reasoning (e.g., morphology, topology, semantic hierarchy) and remain robust despite stylistic variation. Although a large and diverse corpus improves generalization, annotated data remains limited, and geographic or stylistic biases can hinder transfer to truly unseen or underrepresented map styles. To mitigate this, Xia et al.~\cite{xia_contrastive_2023} introduce a contrastive pretraining strategy for Transformers, leveraging image pairs of the same location from different historical map series. To further reduce annotation requirements, in a subsequent work, they introduce MapSAM~\cite{xia_mapsam_2024}, to leverage the strong zero-shot segmentation capabilities of SAM~\cite{kirillov_segment_2023} for historical map segmentation. By integrating domain-specific knowledge via DoRA~\cite{mao_dora_2024} into the image encoder, automating prompt generation, and enhancing both positional prompts and the attention mechanisms, their approach improves effective automatic segmentation of historical maps.

MapSAM is built upon two key pillars of modern computer vision: large-scale, pre-trained vision foundation models, which provide powerful, general-purpose feature representations and the ability to adapt these massive models to new domains efficiently. As mentioned, one prominent example is SAM~\cite{kirillov_segment_2023}, which has demonstrated remarkable generalization capabilities across various segmentation tasks. SAM is optimized for instance-level distinctions to segment clearly defined, prompt-specific objects based on supervised mask annotations~\cite{kirillov_segment_2023}. An alternative to leveraging SAM’s embeddings is the features from DINOv2~\cite{oquab_dinov2_2024}, which learns rich semantic representations by enforcing consistency across different views of the same scene using a self-supervised objective, without relying on labeled data. A model that combines SAM, DINOv2, and other foundation models is RADIO~\cite{ranzinger_am-radio_2024} by Ranzinger et al., who use agglomeration learning to distill multiple foundation models into a single one, yielding even more expressive feature representations.

Concurrently with the development of large vision models, there has also been growing interest in adapting specifically SAM for few-shot segmentation: He et al.~\cite{he_apseg_2024} proposed APSeg with domain-agnostic feature transformation for cross-domain applications. Sun et al.~\cite{sun_vrp-sam_2024} developed VRP-SAM to utilize annotated reference images as segmentation prompts with various annotation formats. Moreover, Liu et al.~\cite{liu2024matcher} as well as Zhang et al.~\cite{zhang2024personalize}, propose similar training-free prototype-based methodologies to extend SAM for few-shot segmentation.

As underscored by Xia et al.~\cite{xia_mapsam_2024} for effectively applying large foundation models, with their general-purpose features, to highly specialized tasks such as historical map segmentation, adaptation is crucial. However, due to the sheer size of these models, often containing billions of parameters, full fine-tuning becomes increasingly impractical. In order to be able to adapt models of this size efficiently, Hu et al.~\cite{hu_lora_2021} introduced LoRA, which initiated a new research direction focused entirely on efficiently fine-tuning large foundation models through low-rank decomposition. Further extensions like DoRA~\cite{mao_dora_2024} improve representational capacity without significantly increasing the number of parameters by decomposing pre-trained weights into magnitude and direction, applying LoRA specifically to the directional component to reduce the number of trainable parameters during fine-tuning.

Apart from these advancements, few-shot segmentation for historical documents has fostered an active research community -- though with a stronger focus on pixel-precise layout analysis of handwritten manuscripts than on historical maps: De Nardin et al.~\cite{denardin2023efficient} propose an efficient learning-based method combined with classical binarization techniques, requiring only two labeled pages per manuscript. Subsequently, they demonstrate that even a single annotated page can suffice when paired with lightweight augmentations and balanced loss functions~\cite{denardin2024one_shot}. Architecture-wise, all three rely on smaller and simpler models such as DeepLabv3+~\cite{chen_deeplab_2017} or its predecessor, rather than adapting large foundation models.

\section{Methodology}

Our methodology follows a three-stage approach to adapt a vision foundation model with only a handful of labels to the specialized domain of historical maps: first, we extract semantically rich image embeddings from a vision foundation model such as SAM~\cite{kirillov_segment_2023}, DINOv2~\cite{oquab_dinov2_2024}, or RADIO~\cite{ranzinger_am-radio_2024}; second, we upscale the embeddings back to the original size and, third, use a linear classifier at the pixel level to obtain logit masks.
Given that these foundation models have been trained predominantly on natural images~\cite{kirillov_segment_2023}, we incorporate low-rank adaptation techniques to enhance predictive performance further.

\subsection{Extracting Image Embeddings}

In this work, we primarily focus on three different foundation models: DINOv2 and RADIO, which have been specifically trained for image feature extraction, as well as SAM, a model that has explicitly been trained for open-vocabulary instance segmentation.
For the latter, we extract solely the image encoder from the segmentation framework and discard the mask decoder head, as opposed to previous work.
All three of these foundation models use the Vision Transformer~(ViT) as their base architecture, which has been introduced by Dosovitskiy et al.~\cite{dosovitskiy_image_2021}, and comes in three different variants: ViT-Base, ViT-Large, and ViT-Huge, with parameter counts ranging from 86 million to 632 million.
While DINOv2 and RADIO use standard ViTs, SAM employs ViTDet~\cite{li_exploring_2022} as a means to reduce computational and memory complexity, which is specific to SAM as it enforces a fixed input size of $1024\times 1024$ pixels to enable pixel-precision segmentation at the cost of requiring more compute.
DINOv2 and RADIO support arbitrary resolutions and aspect ratios~\cite{ranzinger_am-radio_2024}.

Given an input image $\mathbf{x} \in \mathbb{R}^{H \times W \times C}$, the image is first preprocessed with the vision foundation model $\mathcal{F}_\theta$ specific preprocessor and subsequently converted into a sequence of token embeddings, which consists of dividing the image into non-overlapping patches of size $P \times P$, resulting in a sequence of flattened patches:

\begin{equation}
    \mathbf{x}_p \in \mathbb{R}^{N \times (P^2 \times C)},
\end{equation}

where $N = HW / P^2$ is the number of patches~\cite{dosovitskiy_image_2021}.
Next, patches are linearly projected into a $D$-dimensional latent space using a trainable embedding layer:

\begin{equation}
    \mathbf{z}_0 = [\mathbf{x}_{\text{class}}; \mathbf{x}_p^1\mathbf{E}; \mathbf{x}_p^2\mathbf{E}; \dots; \mathbf{x}_p^N\mathbf{E}] + \mathbf{E}_{\text{pos}},
\end{equation}

where $\mathbf{E} \in \mathbb{R}^{(P^2 \times C) \times D}$ is the projection matrix, and $\mathbf{E}_{\text{pos}} \in \mathbb{R}^{(N+1) \times D}$ represents a learnable positional encoding.
Additionally, a special token, $\mathbf{x}_{\text{class}}$, is prepended to the sequence, which serves as a global image embedding.

Next, the token sequence is processed by a Transformer encoder (cf.~\cite{vaswani_attention_2023}) consisting of $L$ layers (24 for ViT-L) of Multi-Head Self-Attention (MSA) and feed-forward networks, with residual connections and layer normalization applied at each step, yielding a final representation: $\mathbf{z}_L$.
At position $\mathbf{z}_L^0$ is the special token, now with global context information.
Although important for downstream tasks such as classification, this token is discarded in our setting, where we require solely spatial features.
After reshaping the output, we end up with strong spatial feature representations in the form of: 

\begin{equation}
    \mathbf{z} = \mathcal{F}_\theta(\mathbf{x}_p), \quad \mathbf{z} \in \mathbb{R}^{H' \times W' \times D},
\end{equation}

where $H' = H / P$ and $W' = W / P$ correspond to the number of patches along each spatial dimension, which can be used to perform segmentation or other downstream tasks such as dense feature matching~\cite{oquab_dinov2_2024}.
When visualizing these embeddings, e.g., by applying principal component analysis and plotting the first three components, clear semantic classes emerge (cf. \Cref{fig:pca_radio}), even for specialized data such as historical maps.

\subsection{Linear-Probing}
Next, keeping the image encoder $\mathcal{F}_\theta$ frozen, we map the embeddings to soft mask prediction.
To map from $\mathbf{z}$ to a soft mask output which can be compared to a given binary annotation mask $\mathbf{m} \in \mathbb{B}^{H \times W}$, we first perform bilinear up-sampling to resize the embeddings back to their original size of $H \times W$ and then perform linear pixel-wise classification via $f$ on each feature vector $\mathbf{z}_{i,j}$:

\begin{equation}
    f(\mathbf{z}_{i,j})= \sigma\Big(\mathbf{z}_{i,j}^T\mathbf{w} + b\Big)=\mathbf{\hat{m}}_{i,j}, \quad \mathbf{w} \in \mathbb{R}^{D}, \quad b \in \mathbb{R},
\end{equation}

where $\mathbf{w}, b$ are learned parameters, and $\sigma(\cdot)$ is the sigmoid function, mapping the logit to a probability.
In our experiments, we found that performing the pixel classification step after the up-sampling is crucial, not to interpolate logits but feature embeddings to obtain more accurate masks.

We explored more complex classifiers, such as the decoder head of the MaskFormer~\cite{cheng_per-pixel_2021} or learned transpose convolutions as well.
However, in our few-shot setting with limited training examples, high-capacity models tended to overfit, resulting in poor generalization. Thus, linear probing emerges as a simple yet effective and robust solution.

\subsection{Fine-Tuning via Low-Rank Adaptation}
Although a prediction can be obtained with the previous two steps, the data domain gap between these vision foundation models might still be significant.
For instance, SAM~\cite{kirillov_segment_2023} has been trained on modern object-centric datasets, so its feature space might not align well with the semantic meaning of cartographic symbols, resulting in inferior performance than what could be achieved if adapted to the domain of historical maps.
However, full-finetuning, which retrains all model parameters, in times of foundation models, becomes impractical due to their enormous parameter count, lying in the order of hundreds of millions for the ViT-L version~\cite{hu_lora_2021}.
According to Kim et al.~\cite{kim_how_2022}, full-finetuning can be disadvantageous as fully fine-tuned feature extractors can distort the rich and strong pre-trained representations.

Hence, we opt for so-called low-rank adaptation (LoRA), a technique developed by Hu et al.~\cite{hu_lora_2021}.
LoRA is a parameter-efficient fine-tuning method that injects trainable low-rank decomposition matrices into pre-trained weight matrices, reducing the number of trainable parameters while maintaining expressiveness.
LoRA is typically applied to the query, key, and value matrices $\mathbf{W}_Q, \mathbf{W}_K, \mathbf{W}_V \in \mathbb{R}^{D\times D_h}$, with $D_h$ being $D$ divided by the number of MSA heads, and the output projection layer $\mathbf{W}_O \in \mathbb{R}^{D\times D}$ of MSA.
Hence, instead of updating, for instance, \( \mathbf{W}_Q \) directly, LoRA reparametrizes it as:

\begin{equation}
    \mathbf{W}_Q^\prime = \mathbf{W}_Q + \mathbf{B} \mathbf{A},
\end{equation}

with $\mathbf{A} \in \mathbb{R}^{{D} \times r}$ and $\mathbf{B} \in \mathbb{R}^{r \times D_h}$, where $r$ is the rank of the decomposition and is typically chosen such that $r \ll \min(D_h, D)$ to reduce the number of trainable parameters significantly.
By setting the rank $r$ to be the same rank as the pre-trained weights, one essential recovers full-finetuning.
We, therefore, select $r=4$ to avoid overfitting with our limited data. 

In addition to LoRA, many other parameter-efficient fine-tuning techniques emerged over the years, including LoKr and LoHa by Hyeon et al.~\cite{hyeon-woo_fedpara_2021}, as well as DoRA~\cite{mao_dora_2024}, which we evaluate in this work.

\subsection{Objective and Optimization}

In order to tune the weights $\mathbf{w}$ and bias $b$ of our probing head $f$, as well as adapting $\mathcal{F}_\theta$ to our task-specific domain of historical maps, we use a mixture of Focal~\cite{lin_focal_2018} and Dice~\cite{sudre_generalised_2017} loss as proposed by Cheng et al.~\cite{cheng_per-pixel_2021}. In detail, our training objective is as follows:

 \begin{equation}
 \mathcal{L} = \alpha\mathcal{L}_{\text{focal}} +\beta \mathcal{L}_{\text{dice}},
 \end{equation}
where $\alpha$ and $\beta$ hyperparameters to weigh the two terms. Compared to Perera et al.~\cite{perera2024discriminative}, we found that $\alpha=10$ and $\beta=1$, a down-weighing of the first term benefited predictive performance.

We optimize the objective by utilizing an AdamW optimizer.
In addition to that, we employ a scheduler to stabilize and expedite training.
Specifically, we opted for the one-cycle policy described in the works by Smith et al.~\cite{smith_super-convergence_2018} which anneals the learning rate from an initial rate ($10^{-4}$) to some maximum ($10^{-3}$) and then from that to some minimum much lower than the initial learning rate.
 
\section{Experiments \& Results}

In the following, we describe the evaluated datasets, a general overview of training and evaluation specifics, as well as the ablations we conducted, which include:
\begin{itemize}
    \item \textbf{Foundation Models:} We compare the performance of DINOv2, RADIO, and SAM as feature extractors, i.e., keeping the encoder frozen and solely training a lightweight classifier.
    \item \textbf{Low-Rank Adaptation:} We assess four different parameter-efficient fine-tuning methods, namely LoRA, DoRA, LoHa, and LoKr, and compare them to pure linear probing of the encoder.
    \item \textbf{Input Resolution:} We examine the model’s performance under different input resolutions, ranging from 224 pixels to the computationally much more intensive resolution of 1120 pixels.
\end{itemize}

We conclude this section with our final results, where we evaluate \textbf{Few-Shot Performance}.
Specifically, we evaluate our best model in a multitude of few-shot settings, spanning $k$ from a single sample to the whole training set.

\subsection{Datasets}
We evaluate our proposed approach on two datasets of historical maps: the \textit{Siegfried}~\cite{xia_mapsam_2024} dataset and the \textit{ICDAR 2021}~\cite{chazalon_icdar_2021} dataset, specifically on the task of segmenting building blocks from non-building blocks.

\subsubsection{Siegfried} consists of two sub-datasets: a railway dataset representing linear features and a vineyard dataset representing areal features.
Both datasets are divided into $224\times224$ pixel patches from the Swiss Siegfried Maps\footnote{see \href{https://www.swisstopo.admin.ch/en/siegfried-map}{https://www.swisstopo.admin.ch/en/siegfried-map}, last access: 15.03.2025.}.
In total, the full railway dataset consists of 5,872 training tiles, 839 validation tiles, and 1,679 testing tiles, maintaining an approximate 7:1:2 split.
When examining predictive performance in few-shot settings, we systematically reduce the number of labeled training tiles from 100\% to 10\% (587 tiles), 1\% (58 tiles), 10 tiles, 5 tiles, and 1 tile.
In the same fashion, the much smaller vineyard dataset is split into 613 training tiles, 87 validation tiles, and 177 testing tiles.
Here, based on the number of available samples, we solely conduct training with 100\%, 10-shot, 5-shot, and 1-shot experiments.
Finally, overall performance is evaluated on the test set in each scenario.

\subsubsection{ICDAR 2021} comprises historical maps of Paris (1860s–1940s) with the task of detecting closed shapes representing building blocks, which are separated by streets, rivers, or fortifications.
Concerning technical details, the dataset includes large-scale scans (up to $8000\times8000$ pixels) with a frame mask distinguishing relevant from non-relevant areas. For evaluation, we crop, for simplicity, non-overlapping $448\times448$ patches, resize them to $224\times224$, and rescale predictions accordingly.
In total, it consists of one training image, one validation image, and three test images.

\subsection{Experimental Setup}

We report the following standard evaluation metrics for semantic segmentation: the mean Intersection over Union (mIoU) as well as the F1-Score (F1).
To ensure the validity of our results, we trained a minimum of three models and averaged their results.
In general, our ablations are performed on the Siegfried dataset, using the same ten samples as by Xia et al.~\cite{xia_mapsam_2024} and training for 300 epochs with a batch size of four.
With respect to our three image encoders, we use the ViT-L variants.
While training, we track the mIoU of the validation set, which we use to select the model for the final evaluation on the test set.

To increase data variety in our few-shot setting, we employ $D_4$-dihedral group symmetry transformations.
These transformations include \textit{rotations} (90°, 180°, 270°), \textit{flips} (vertical, horizontal, diagonal), and \textit{transpositions}.
A great benefit of these augmentations is that input data remains within its original data distribution.
We also considered more drastic augmentations in the form of MixUp~\cite{zhang_mixup_2018} and CutMix~\cite{yun_cutmix_2019}. 
However, we found this to be not beneficial in our case, which is in accordance with the findings by Kim et al.~\cite{kim_how_2022}, who found out that augmentations with strong intensity degrade performance in few-shot learning.

We pursue a similar yet less drastic approach to input resolution than SAM.
To obtain spatially larger feature embeddings, we rescale the input image by default to three times its size ($3\cdot224$ pixels), to improve per-pixel accuracy.

Concerning hardware, our experiments were conducted on NVIDIA RTX A5000/A6000 GPUs, depending on the required memory.
More specifically, evaluations using the image encoder from SAM or evaluating input resolutions greater than $3 \cdot 224)$ pixels required us to use GPUs with bigger memory and reduce the batch size down to only two samples.

\subsection{Ablation Study}

In the following sections, we report the results of our conducted ablations.
As a teaser, we find that among the three foundation models, when probing them for segmentation, RADIO has the richest feature embeddings. We find DoRA and LoRA to perform very similarly for parameter-efficient fine-tuning, but opted for DoRA due to its generally strong reported performance~\cite{xia_mapsam_2024}.
Finally, we find up-scaling the input beneficial, with three times the original input resolution performing best.

\begin{figure}[t]
\centering
\begin{subfigure}{0.24\textwidth}
\includegraphics[width=\textwidth]{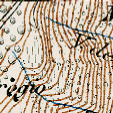}
\subcaption{Input}
\end{subfigure}
\begin{subfigure}{0.24\textwidth}
\includegraphics[width=\textwidth]{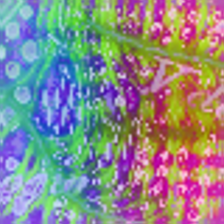}
\subcaption{DINOv2}
\end{subfigure}
\begin{subfigure}{0.24\textwidth}
\includegraphics[width=\textwidth]{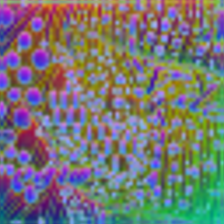}
\subcaption{SAM}
\end{subfigure}
\begin{subfigure}{0.24\textwidth}
\includegraphics[width=\textwidth]{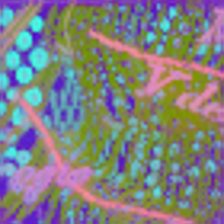}
\subcaption{RADIO}
\end{subfigure}
\caption{A visualization of the first three principal components of feature embeddings from vision foundation models: subjectively speaking, RADIO exhibits the strongest spatial features, followed by DINOv2 and, lastly, SAM.}
\label{fig:pca_comparison}
\end{figure}

\subsubsection{Foundation Models.}

In the following, we quantitatively and qualitatively assess DINOv2, RADIO, and SAM when keeping their weights fixed and training a linear classifier to predict segmentation masks per-pixel.

\paragraph{Qualitative Results.} Similar to the works by Oquab et al.~\cite{oquab_dinov2_2024}, we use PCA to reduce the spatial feature dimensionality of patch features from the three foundation models and visualize their first three principal components.
\Cref{fig:pca_comparison} illustrates the ability of the selected foundation models to separate historical maps semantically into different classes.
In all three cases, classes are separated; however, subjectively speaking, RADIO exhibits the strongest spatial features, followed by DINOv2 and, lastly, SAM.
RADIO clearly distinguishes between text, contour lines, empty land, and vegetation symbols.
Opposed to this is SAM, which puts text, vegetation symbols, and empty land semantically closer.

As SAM is optimized for instance-level segmentation to distinct clearly defined, prompt-specific objects based on supervised mask annotations, this is to be expected~\cite{kirillov_segment_2023}.
In contrast, DINOv2 is trained with a self-supervised objective that encourages consistency across different views of the same scene, without relying on labeled data~\cite{oquab_dinov2_2024}.
Based on these fundamentally different training goals, it is clear why DINOv2 -- and by extension, RADIO through its agglomerative learning -- offers richer and more general representations.

\begin{table}
\centering
\caption{Comparison of segmentation performance for DINOv2, SAM, and RADIO under linear probing with frozen backbones.}
\begin{tabular}{l c cc cc}
\toprule
        \multirow{2}{*}{\textbf{Method}} & & \multicolumn{2}{c}{\textbf{Railways}} & \multicolumn{2}{c}{\textbf{Vineyards}}\\
        & & F1 & IoU & F1 & IoU \\
\midrule
DINOv2 & \cite{oquab_dinov2_2024} & 35.5 & 19.1 & 41.3 & 23.9 \\
SAM & \cite{kirillov_segment_2023} & 45.5 & 16.4 & 44.2 & \textbf{47.4} \\
RADIO & \cite{ranzinger_am-radio_2024} & \textbf{73.4} & \textbf{52.4} & \textbf{62.1} & 36.1 \\
\bottomrule
\end{tabular}
\label{tab:foundation_models}
\end{table}

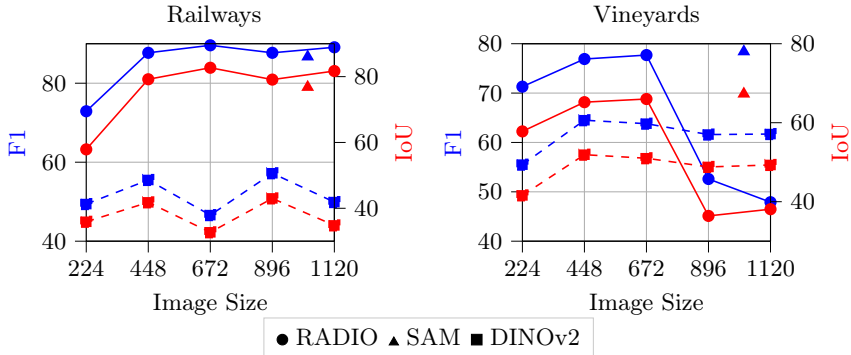
\begin{figure}
    \centering
    \begin{tikzpicture}

\definecolor{darkgray176}{RGB}{176,176,176}
\definecolor{lightgray204}{RGB}{204,204,204}
\definecolor{orange}{RGB}{255,165,0}

\begin{axis}[
legend style={empty legend},
name=ax1,
tick align=outside,
tick pos=left,
x grid style={darkgray176},
xmajorgrids,
xmin=1, xmax=5,
width=0.4\textwidth,
xtick style={color=black},
%xtick={0.05,0.2,0.4,0.6,0.8,0.95}, % definitions of ytick labels (left)
xticklabels={~, 224, 448, 672, 896, 1120},
y grid style={darkgray176},
ylabel=\textcolor{blue}{F1},
ymajorgrids,
ymin=40, ymax=90,
ytick style={color=black},
xlabel={Image Size} 
]
\addplot [semithick, mark=*, blue]
table {%
1 72.9
2 87.7
3 89.6
4 87.7
5 89.1
};\label{plot_one}

\addplot[semithick, dashed, mark=square*, blue]
table {%
1 49.4
2 55.5
3 46.6
4 57.2
5 49.9
};

% \addplot [semithick, blue]
% table {%
% 0.05 55.9757649898529
% 0.1 56.3466191291809
% 0.15 56.5863847732544
% 0.2 56.7810535430908
% 0.25 56.9929659366608
% 0.3 57.1075141429901
% 0.35 57.1800589561462
% 0.4 57.2434306144714
% 0.4 57.3032438755035
% 0.5 57.3366165161133
% 0.55 57.3895931243896
% 0.6 57.4462234973907
% 0.65 57.4805200099945
% 0.7 57.5565218925476
% 0.75 57.4385702610016
% 0.8 57.3426723480225
% 0.85 57.2483003139496
% 0.9 57.1866512298584
% 0.95 57.0111393928528
% };\label{plot_two}
\addplot [blue, mark = triangle*, thick,every node near coord/.style={anchor=north}] coordinates {( 4.57, 86.7)}; % F1
\end{axis}

\begin{axis}[
axis y line=right,
width=0.4\textwidth,
legend cell align={left},
legend style={
  empty legend,
  fill opacity=0.8,
  draw opacity=1,
  text opacity=1,
  at={(1.38,-0.6)},
  anchor=south,
  draw=lightgray204
},
legend columns=3, 
tick align=outside,
axis line style={-},
x grid style={darkgray176},
xmin=1, xmax=5,
xtick style={draw=none},
xticklabels={},
%xtick pos=left,
%xtick style={color=black},
%yticklabel style={anchor=west}, % definitions of ytick labels (right) and xticklabels
y grid style={darkgray176},
ylabel=\textcolor{red}{IoU},
ymin=30, ymax=90,
ytick pos=right,
ytick style={color=black},
yticklabel style={anchor=west}]

%\addlegendimage{/pgfplots/refstyle=plot_two}\addlegendentry{FP removed}
%\addlegendimage{/pgfplots/refstyle=plot_one}\addlegendentry{FP not removed} % legend entries

\addplot [semithick, red, mark=*, forget plot]
table {%
1 57.9
2 79.2
3 82.7
4 79.1
5 81.7
};

\addplot[semithick, dashed, mark=square*, red]
table {%
1 35.9
2 41.8
3 32.7
4 43.0
5 34.8
};

% \addplot [semithick, orange, forget plot]
% table {%
% 0.05 45.1001733541489
% 0.1 45.2396720647812
% 0.15 45.2828526496887
% 0.2 45.371288061142
% 0.25 45.4859048128128
% 0.3 45.5652296543121
% 0.35 45.5559253692627
% 0.4 45.5189526081085
% 0.4 45.49860060215
% 0.5 45.4695105552673
% 0.55 45.4621940851212
% 0.6 45.3925132751465
% 0.65 45.3452855348587
% 0.7 45.271560549736
% 0.75 45.1484233140945
% 0.8 44.8718547821045
% 0.85 44.6612983942032
% 0.9 44.4061487913132
% 0.95 43.9666509628296
% };
\addplot [red, mark = triangle*, thick, every node near coord/.style={anchor=north}] coordinates {( 4.57, 76.8)}; % IOU

\addlegendentry{\myplotmark{*} RADIO }
\addlegendentry{\myplotmark{triangle*} SAM}
\addlegendentry{\myplotmark{square*} DINOv2}
\addlegendimage{}

% \addplot [orange, mark = x, nodes near coords=SAM,every node near coord/.style={anchor=north}] coordinates {( 4.57, 85)};

\end{axis}

%%%%%%%%%%%%%%%%%
\begin{axis}[
tick align=outside,
width=0.4\textwidth,
at={(ax1.south east)},
xshift=2.5cm,
tick pos=left,
x grid style={darkgray176},
xmajorgrids,
xmin=1, xmax=5,
xtick style={color=black},
xticklabels={~, 224, 448, 672, 896, 1120},
%xtick={0.05,0.2,0.4,0.6,0.8,0.95}, % definitions of ytick labels (left)
%xticklabels={0.05,0.2,0.4,0.6,0.8,0.95},
y grid style={darkgray176},
ylabel=\textcolor{blue}{F1},
ymajorgrids,
ymin=40, ymax=80,
ytick style={color=black},
xlabel={Image Size} 
]
\addplot [semithick, mark=*, blue]
table {%
1 71.3
2 76.9
3 77.7
4 52.6
5 47.9
};\label{plot_one}

\addplot[semithick, dashed, mark=square*, blue]
table {%
1 55.5
2 64.5
3 63.8
4 61.6
5 61.7
};

% \addplot [semithick, blue]
% table {%
% 0.05 55.9757649898529
% 0.1 56.3466191291809
% 0.15 56.5863847732544
% 0.2 56.7810535430908
% 0.25 56.9929659366608
% 0.3 57.1075141429901
% 0.35 57.1800589561462
% 0.4 57.2434306144714
% 0.4 57.3032438755035
% 0.5 57.3366165161133
% 0.55 57.3895931243896
% 0.6 57.4462234973907
% 0.65 57.4805200099945
% 0.7 57.5565218925476
% 0.75 57.4385702610016
% 0.8 57.3426723480225
% 0.85 57.2483003139496
% 0.9 57.1866512298584
% 0.95 57.0111393928528
% };\label{plot_two}
\addplot [blue, mark = triangle*, thick,every node near coord/.style={anchor=north}] coordinates {( 4.57, 78.4)}; % F1

\end{axis}

\begin{axis}[
axis y line=right,
at={(ax1.south east)},
xshift=2.5cm,
width=0.4\textwidth,
legend cell align={left},
legend style={
  fill opacity=0.8,
  draw opacity=1,
  text opacity=1,
  at={(0.09,0.5)},
  anchor=west,
  draw=lightgray204
},
tick align=outside,
axis line style={-},
x grid style={darkgray176},
xmin=1, xmax=5,
xtick style={draw=none},
xticklabels={},
%xtick pos=left,
%xtick style={color=black},
%yticklabel style={anchor=west}, % definitions of ytick labels (right) and xticklabels
y grid style={darkgray176},
ylabel=\textcolor{red}{IoU},
ymin=30, ymax=80,
ytick pos=right,
ytick style={color=black},
yticklabel style={anchor=west},
]

%\addlegendimage{/pgfplots/refstyle=plot_two}\addlegendentry{FP removed}
%\addlegendimage{/pgfplots/refstyle=plot_one}\addlegendentry{FP not removed} % legend entries

\addplot [semithick, red, mark=*, forget plot]
table {%
1 57.8
2 65.2
3 66.0
4 36.4
5 38.1
};
\addplot[semithick, dashed, mark=square*, red]
table {%
1 41.6
2 51.9
3 51.0
4 48.8
5 49.3
};
% \addplot [semithick, orange, forget plot]
% table {%
% 0.05 45.1001733541489
% 0.1 45.2396720647812
% 0.15 45.2828526496887
% 0.2 45.371288061142
% 0.25 45.4859048128128
% 0.3 45.5652296543121
% 0.35 45.5559253692627
% 0.4 45.5189526081085
% 0.4 45.49860060215
% 0.5 45.4695105552673
% 0.55 45.4621940851212
% 0.6 45.3925132751465
% 0.65 45.3452855348587
% 0.7 45.271560549736
% 0.75 45.1484233140945
% 0.8 44.8718547821045
% 0.85 44.6612983942032
% 0.9 44.4061487913132
% 0.95 43.9666509628296
% };
\addplot [red, mark = triangle*, thick, every node near coord/.style={anchor=north}] coordinates {( 4.57, 67.3)}; %IoU
\end{axis}

\node at (1.7,3) {Railways};
\node at (7.45,3) {Vineyards};

\end{tikzpicture}
    \caption{Analyzing the impact of resolution: RADIO is on par with / better than SAM, while being computationally more efficient~\cite{ranzinger_am-radio_2024}.}\label{fig:size_ablation}    
\end{figure}

\paragraph{Quantitative Results.}
Next, we examine the three foundation models quantitatively with the results denoted in~\Cref{tab:foundation_models}.
Continuing with the assumption that RADIO offers the most expressive features, the results are not as definitive: for the railway dataset, RADIO performs best by a large margin (mIoU of 52) followed by DINOv2 (mIoU of 19); in contrast, for the vineyard dataset SAM performs best (mIoU of 47) followed by RADIO (mIoU of 36).
Although ambiguous, we show in subsequent sections that RADIO is equal to or better than SAM~(cf.~\Cref{fig:size_ablation}), with the additional benefit of requiring less memory.

\begin{table}
\centering
\caption{Comparison of parameter-efficient fine-tuning methods by segmentation performance and number of trainable parameters.}
\begin{tabular}{l c cc cc | r c}
\toprule
        \multirow{2}{*}{\textbf{Method}} & & \multicolumn{2}{c}{\textbf{Railways}} & \multicolumn{2}{c}{\textbf{Vineyards}} & \multicolumn{2}{c}{\textbf{Parameters}}\\
      & & F1 & IoU & F1 & IoU & Total & in \% \\
\midrule
None & & 73.4 & 52.4 & 62.1 & 36.1 & 1k & 0.00\% \\
LoRA & \cite{hu_lora_2021} & 89.6 & 82.5 & \textbf{78.6} & \textbf{67.9} & 590k & 0.18\% \\
LoKr & \cite{hyeon-woo_fedpara_2021} & 87.1 & 78.8 & 76.1 & 63.7 & 77k & 0.02\% \\
LoHa & \cite{hyeon-woo_fedpara_2021} & 88.1 & 80.2 & 77.9 & 66.8 & 1,180k & 0.37\% \\
DoRA & \cite{mao_dora_2024} & \textbf{89.9} & \textbf{83.7} & 77.8 & 67.2 & 689k & 0.21\% \\
\bottomrule
\end{tabular}
\label{tab:peft}
\end{table}

\subsubsection{Low-Rank Adaptation.}
In the following, we evaluated four different low-rank adaptation methods: LoRA, LoKr, LoHa, and DoRA.
We define parameter-efficient fine-tuning configurations tailored to different data regimes.
For few-shot learning with limited samples ($k\leq10$), we employed a lower rank ($r=4$), moderate scaling factor ($\alpha=8$), and higher dropout rate (20\%) to mitigate overfitting.
Conversely, for datasets with more samples ($>10$), we increased the rank ($r=8$) and scaling factor ($\alpha=16$) while reducing the dropout rate to 10\%, improving learning stability and reducing the risk of overfitting.

Considering the results shown in \Cref{tab:peft}, we find that simply using the embeddings as is and probing the network for semantic segmentation yields the worst performance.
In contrast, adapting the model via low-rank adaptation shows significant improvements, with either LoRA or DoRA performing best depending on the dataset.
Our results match previous findings, as according to Kim et al.~\cite{kim_how_2022} fine-tuning is a better strategy than linear probing.

Although almost indifferent, we selected DoRA for all subsequent experiments to allow for a better comparison to MapSAM~\cite{xia_mapsam_2024}.
Notably, implementing parameter-efficient fine-tuning for the RADIO-L variant required only an additional 689k parameters, representing just 0.21\% of the total trainable parameters.

\subsubsection{Input Scaling.}
In the next ablation, we consider the impact of input resolution and report the median over three runs due to strong variations in performance exhibited.
Ablating input size plays a crucial role, as larger inputs result in higher-dimensional feature embeddings from the encoder.
For instance, in SAM, resizing the input to a resolution of $1024\times1024$ pixels is the default to benefit precision in segmentation~\cite{kirillov_segment_2023}.
However, improved per-pixel performance comes at the cost of requiring much more compute, with the MSA being the bottleneck as its complexity is in $O(N^2)$, i.e., compute grows quadratically w.r.t.\ the number of patches.
In our experiments, SAM took about five times longer than RADIO with an input resolution of $3\cdot(224\times224)$.

We summarize our findings in \Cref{fig:size_ablation}, which depicts the performance of the three foundation models, adapted using DoRA~\cite{mao_dora_2024} at different input resolutions.
As SAM operates at a fixed input resolution, we report its performance at 1024 pixels.
Analyzing the two graphs shows that RADIO peaks at a resolution of 672 pixels and performs on par with / better than SAM while being computationally cheaper and requiring less memory.
Although this performance remains constant for the railway dataset, we experience a drastic drop in performance for vineyards with higher pixel values.
We assume that this is due to the mode switch reported by Ranzinger et al.~\cite{ranzinger_am-radio_2024}, where performance drops at a resolution greater than 720px; however, the performance on railways seems not to be impacted by this.
Generally speaking, we experienced fluctuations in vineyards for both RADIO and SAM (not for DINOv2), and we hypothesize that sample quality is more important than for railways.
Based on these insights, we keep the input resolution at 672 pixels and proceed with RADIO for the final evaluation.

\subsection{Results}
In the final evaluation, we compare our method to existing approaches on the Siegfried and ICDAR 2021 datasets. Our method, a RADIO-L model, low-rank adapted with DoRA and probed with a linear classifier for per-pixel segmentation, achieves competitive results despite its simplicity: it outperforms the SOTA in all few-shot settings on the Siegfried dataset for both railways and vineyards.

\begin{figure}[t]
\centering
\begin{subfigure}{0.16\textwidth}
\includegraphics[width=\textwidth]{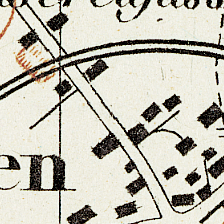}
\end{subfigure}
\begin{subfigure}{0.16\textwidth}
\includegraphics[width=\textwidth]{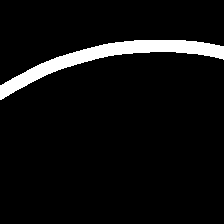}
\end{subfigure}
\begin{subfigure}{0.16\textwidth}
\includegraphics[width=\textwidth]{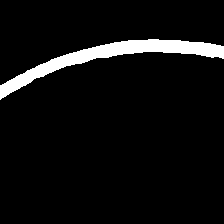}
\end{subfigure}
\begin{subfigure}{0.16\textwidth}
\includegraphics[width=\textwidth]{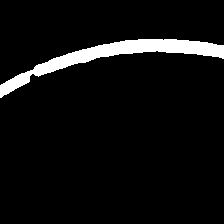}
\end{subfigure}
\begin{subfigure}{0.16\textwidth}
\includegraphics[width=\textwidth]{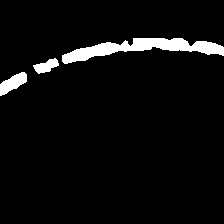}
\end{subfigure}

\begin{subfigure}{0.16\textwidth}
\includegraphics[width=\textwidth]{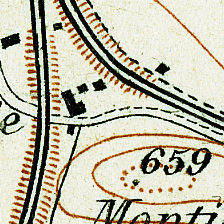}
\end{subfigure}
\begin{subfigure}{0.16\textwidth}
\includegraphics[width=\textwidth]{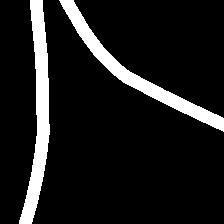}
\end{subfigure}
\begin{subfigure}{0.16\textwidth}
\includegraphics[width=\textwidth]{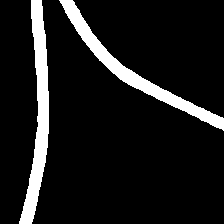}
\end{subfigure}
\begin{subfigure}{0.16\textwidth}
\includegraphics[width=\textwidth]{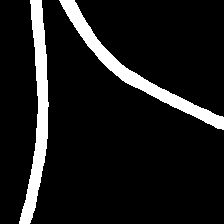}
\end{subfigure}
\begin{subfigure}{0.16\textwidth}
\includegraphics[width=\textwidth]{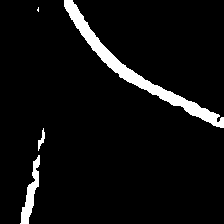}
\end{subfigure}

\begin{subfigure}{0.16\textwidth}
\includegraphics[width=\textwidth]{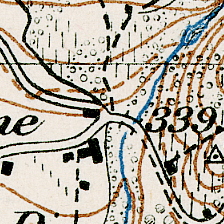}
\end{subfigure}
\begin{subfigure}{0.16\textwidth}
\includegraphics[width=\textwidth]{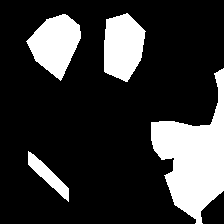}
\end{subfigure}
\begin{subfigure}{0.16\textwidth}
\includegraphics[width=\textwidth]{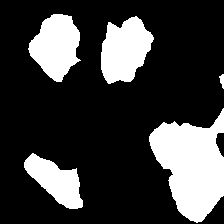}
\end{subfigure}
\begin{subfigure}{0.16\textwidth}
\includegraphics[width=\textwidth]{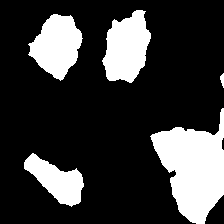}
\end{subfigure}
\begin{subfigure}{0.16\textwidth}
\includegraphics[width=\textwidth]{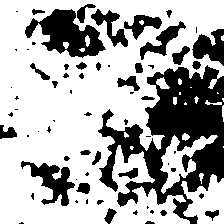}
\end{subfigure}

\begin{subfigure}{0.16\textwidth}
\includegraphics[width=\textwidth]{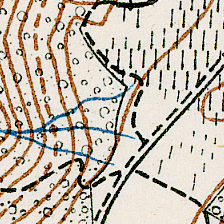}
\subcaption{Input}\label{fig:input_ex}
\end{subfigure}
\begin{subfigure}{0.16\textwidth}
\includegraphics[width=\textwidth]{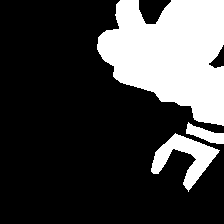}
\subcaption{Label}\label{fig:gt_ex}
\end{subfigure}
\begin{subfigure}{0.16\textwidth}
\includegraphics[width=\textwidth]{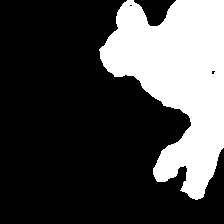}
\subcaption{10-Shot}\label{fig:10shot_ex}
\end{subfigure}
\begin{subfigure}{0.16\textwidth}
\includegraphics[width=\textwidth]{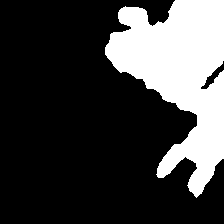}
\subcaption{5-Shot}\label{fig:5shot_ex}
\end{subfigure}
\begin{subfigure}{0.16\textwidth}
\includegraphics[width=\textwidth]{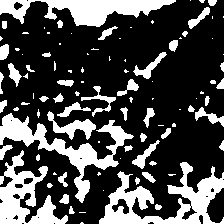}
\subcaption{1-Shot}\label{fig:1shot_ex}
\end{subfigure}
\caption{Examples of the Siegfried dataset~\cite{xia_mapsam_2024}, showing (\subref{fig:input_ex}) the input map, (\subref{fig:gt_ex}) the ground truth, and the predictions at (\subref{fig:10shot_ex}) 10-shots, (\subref{fig:5shot_ex}) 5-shots, and (\subref{fig:1shot_ex}) 1-shot.}
\label{fig:siegfried_results}
\end{figure}

\begin{table}
    \centering
    \caption{Comparing the segmentation performance of SAM variants against a U-Net and our approach, based on RADIO, on the railway dataset.}
    \renewcommand{\arraystretch}{1.2}
    \begin{tabular}{lcc cc cc cc cc cc}
        \toprule
        \multicolumn{13}{c}{\textbf{Railways}}\\
        \midrule
        \multirow{2}{*}{\textbf{Method}} & \multicolumn{2}{c}{Full (5872)} & \multicolumn{2}{c}{10\% (587)} & \multicolumn{2}{c}{1\% (58)} & \multicolumn{2}{c}{10-shot} & \multicolumn{2}{c}{5-shot} & \multicolumn{2}{c}{1-shot}\\
        & F1 & IoU & F1 & IoU & F1 & IoU & F1 & IoU & F1 & IoU & F1 & IoU \\
        \midrule
        U-Net~\cite{ronneberger_u-net_2015} & 96.2 & 92.1 & 94.9 & 90.2 & 86.6 & 72.0 & 82.2 & 70.5 & 72.3 & 54.7 & 14.4 & 9.5 \\
        DeepLabV3+~\cite{chen_encoder-decoder_2018} & 96.6 & 92.9 & 95.6 & 91.0 & 92.8 & 85.0 & 83.2 & 72.2 & 75.8 & 62.1 & 28.1 & 16.4 \\
        SegFormer~\cite{xie_segformer_2021} & 96.6 & \textbf{93.0} & 96.0 & 91.8 & 94.1 & 87.6 & 87.8 & 78.0 & 81.5 & 69.3 & 36.4 & 23.8 \\
            SAMed\footnotemark[1] & 92.0 & 86.3 & 91.6 & 85.7 & 91.8 & 86.0 & 84.6 & 75.4 & - & - & - & -\\
%        Few-Shot SAM\footnotemark[1] & -- & -- & -- & -- & -- & -- & 47.5 & 35.8 & - & - & - & -\\
        MapSAM & 94.1\footnotemark[1] & 89.5\footnotemark[1] & 93.6\footnotemark[1] & 88.7\footnotemark[1] & 92.1\footnotemark[1] & 86.5\footnotemark[1] & 87.2\footnotemark[1] & 78.5\footnotemark[1] & 75.2 & 64.3 & 40.3 & 34.6\\
        \midrule
        Ours & \textbf{96.7} & 92.0 & \textbf{96.4} & \textbf{92.5} & \textbf{94.9} & \textbf{90.5} & \textbf{89.4} & \textbf{82.0} & \textbf{86.1} & \textbf{77.0} & \textbf{52.4} & \textbf{41.5}  \\
        \bottomrule
    \end{tabular}
    \label{tab:railway_results}
\end{table}

\subsubsection{Siegfried}
In this section, we evaluate our model on the Siegfried dataset against other baselines and MapSAM~\cite{xia_mapsam_2024}.
For this, we train a baseline U-Net~\cite{ronneberger_u-net_2015} with a ResNet-50~\cite{he2016deep} encoder for 100 epochs.
In addition, we evaluate two other model variants: DeepLabV3+~\cite{chen_deeplab_2017} \& SegFormer~\cite{xie_segformer_2021}.
We selected the former as it poses a strong CNN-based baseline, which was additionally employed successfully in the context of few-shot segmentation of historical documents~\cite {denardin2024one_shot,denardin2023efficient}.
For the latter, we wanted to contrast CNN baselines with a transformer-based architecture.
With respect to our model, as more data is available, i.e., when training with 100\% and 10\%, the number of epochs is reduced to 30, and for runs with very few examples, i.e., $k\leq5$, results are averaged over 10 runs.

First, we evaluate our model on railways with its results denoted in \Cref{tab:railway_results}, which indicates the strong performance of our simple yet effective approach as it outperforms the SOTA in all few-shot settings.
Notably, it achieves a +5\% improvement in the 10-shot setting and a 20\% gain in the more challenging 5-shot setting.
Second, our approach also yields the best performance in all few-shot settings for the vineyard dataset.
Analyzing the results in \Cref{tab:vineyard_results}, we observe an improvement of +13\% in the 10-shot setting and +23\% in the 5-shot setting.

\begin{table}
    \centering
    \caption{Comparing the segmentation performance of SAM variants against a U-Net and our approach, based on RADIO, on the vineyard dataset.}
    \renewcommand{\arraystretch}{1.2}
    \begin{tabular}{lcc cc cc cc}
        \toprule
        \multicolumn{9}{c}{\textbf{Vineyards}}\\
        \midrule
        \multirow{2}{*}{\textbf{Method}} & \multicolumn{2}{c}{Full (613)} & \multicolumn{2}{c}{10-shot} & \multicolumn{2}{c}{5-shot} & \multicolumn{2}{c}{1-shot}\\
        & F1 & IoU & F1 & IoU & F1 & IoU & F1 & IoU\\
        \midrule
        U-Net~\cite{ronneberger_u-net_2015} & 80.3 & 69.3 & 56.0 & 39.5 & 51.3 & 36.5 & 29.3 & 18.6  \\
        DeepLabV3+~\cite{chen_encoder-decoder_2018} & 83.9 & 72.9 & 67.9 & 52.7 & 62.3 & 44.1 & 34.4 & 21.4 \\
        SegFormer~\cite{xie_segformer_2021} & 83.2 & 73.7 & 76.3 & 63.5 & 72.7 & 58.7 & 35.5 & 23.9 \\
        SAMed\footnotemark[1] & 82.8 & \textbf{74.9} & 72.0 & 61.5 & - & - & - & - \\
%        Few-Shot SAM\footnotemark[1] & - & - & 58.0 & 46.8 & - & - & - & - \\
        MapSAM & 82.8\footnotemark[1] & 74.3\footnotemark[1] & 70.5\footnotemark[1] & 60.0\footnotemark[1] & 64.7 & 51.6 & 32.9 & 21.9 \\
        \midrule
        Ours & \textbf{86.2} & 74.4 & \textbf{78.5} & \textbf{67.9} & \textbf{75.2} & \textbf{63.3} & \textbf{47.6} & \textbf{34.6} \\
        \bottomrule
    \end{tabular}
    \label{tab:vineyard_results}
\end{table}

\footnotetext[1]{results are taken from \cite{xia_mapsam_2024}.}

\subsubsection{ICDAR2021}
For this dataset, not only are the F1 and mIoU computed.
According to the competition details, instance segmentation performance is evaluated using the COCO Panoptic Quality ($PQ$) metric (cf.~\cite{chazalon_icdar_2021}): first, connected components are extracted from the predicted label mask, from which Segmentation Quality ($SQ$), the mIoU of matched regions, and Recognition Quality ($RQ$), the F-score of detected regions, is computed; the final score is defined as:
\begin{equation}
PQ = SQ \times RQ,
\end{equation}
where a predicted shape is considered a match if its IoU with a reference shape exceeds 0.5; higher is better.
Our performance on the first task of the ICDAR 2021 competition is denoted in \Cref{tab:icdar_results}.
Our simple approach yields an average $PQ$ of 67.3\%, placing us \textit{second} in the rankings.
In terms of IoU and F1-Score, we achieve 77.1 for the former and 82.7 for the latter on the test set.

We find this result striking, given that our model has not been tuned in any way to the shape-aware evaluation criteria of the $PQ$, exemplifying the generalization ability of our methodology.

%\begin{table}[h]
%    \centering
%    \caption{Our results on the ICDAR 2021 challenge about building block segmentation.}
%    \begin{tabular}{ll rrr}
%    \toprule
%    \textbf{Map} & \textbf{Method} & \textbf{PQ} & \textbf{SQ} & \textbf{RQ}\\
%    \midrule
%    301 & \multirow{3}{*}{Ours} & 71.3 & 93.3 & 76.4 \\
%    302 & & 64.2 & 93.2 & 68.9 \\
%    303 & & 66.4 & 93.4 & 71.1 \\
%    \midrule
%    \multirow{5}{*}{\textbf{Mean}} & $1^\text{st}$@ICDAR & 74.1 & - & - \\
%    & Ours & 67.3 & 93.3 & 72.1 \\
%    & $2^\text{nd}$@ICDAR& 62.6 & - & - \\
%    & $3^\text{rd}$@ICDAR& 44.0 & - & - \\
%    & UNet & 14.4 & 86.7 & 16.4 \\
%    \bottomrule
%    \end{tabular}
%    \label{tab:icdar_results}
%\end{table}

\begin{table}
\centering
\caption{Our results on the ICDAR 2021 challenge on building segmentation.}
\begin{minipage}{0.48\textwidth}
    \centering
    \textbf{Per-Map Results} \\
    \vspace{0.2em}
    \begin{tabular}{lrrr}
        \toprule
        \textbf{Map} & \textbf{PQ} & \textbf{SQ} & \textbf{RQ} \\
        \midrule
        301 & 71.3 & 93.3 & 76.4 \\
        302 & 64.2 & 93.2 & 68.9 \\
        303 & 66.4 & 93.4 & 71.1 \\
        \midrule
        Ours & 67.3 & 93.3 & 72.1 \\
        \bottomrule
    \end{tabular}
\end{minipage}
\hfill
\begin{minipage}{0.48\textwidth}
    \centering
    \textbf{Mean Comparisons} \\
        \vspace{0.2em}
    \begin{tabular}{lrc}
        \toprule
        \textbf{Method} & \textbf{PQ} \\
        \midrule
        $1^\text{st}$ Place & 74.1 \\
        \textbf{Ours} & \textbf{67.3} \\
        $2^\text{nd}$ Place & 62.6 \\
        $3^\text{rd}$ Place & 44.0 \\
        U-Net~\cite{ronneberger_u-net_2015} & 14.4 \\
        \bottomrule
    \end{tabular}
\end{minipage}
\label{tab:icdar_results}
\end{table}

\vspace{-0.3cm}
\section{Conclusion}
In this work, we addressed the challenges posed by the diverse visual representations and limited annotated data of historical maps, which are rich sources of historical information. We proposed a simple yet effective approach for few-shot segmentation of historical maps, leveraging the rich semantic embeddings of large vision foundation models combined with parameter-efficient fine-tuning. Our method outperforms the state-of-the-art on the Siegfried benchmark dataset in vineyard and railway segmentation, achieving +5\% and +13\% relative improvements in mIoU in 10-shot scenarios and around +20\% in the more challenging 5-shot setting. Additionally, it demonstrates strong performance on the ICDAR 2021 competition dataset, attaining a mean PQ of 67.3\% for building block segmentation, despite not being optimized for this shape-sensitive metric, underscoring its generalizability. Notably, our approach maintains high performance even in extremely low-data regimes while requiring only 689k trainable parameters -- just 0.21\% of the total model size. In summary, our approach enables precise segmentation of diverse historical maps while drastically reducing the need for manual annotations, advancing automated processing and analysis in the field. Future work could explore test-time fine-tuning with the nearest neighbor example, as suggested by Sun et al.~\cite{hardt2024testtime}, and incorporate test-time augmentation techniques proposed by Kim et al.~\cite{kim_how_2022}.

\bibliographystyle{splncs04}
\bibliography{references}

\end{document}